\documentclass{article}
\usepackage{spconf,amsmath,graphicx,hyperref}
\usepackage{amssymb,booktabs,tikz}
\usepackage[expansion=false]{microtype}
\newcommand{\stale}{\mathrm{stale}}
\newcommand{\cancel}{\mathrm{cancel}}

\newcommand{\WAIT}{\textsc{Wait}}
\newcommand{\COMMIT}{\textsc{Commit}}
\newcommand{\CANCEL}{\textsc{Cancel}}

\title{RASSG: Repair-Aware Semantic Survival Gating for Streaming Tool Execution}

\twoauthors
  {Daijin Hu}
  {Beihang University}
  {Xi Wang}
  {Beihang University}

\begin{document}
\ninept
\renewcommand{\baselinestretch}{1}
\fontsize{9}{9.2}\selectfont
\setlength{\emergencystretch}{3em}

\maketitle
\begin{abstract}
Streaming speech assistants can launch tools before users finish speaking, overlapping execution with incoming speech to reduce user waiting time. However, subsequent self-corrections can invalidate apparently stable arguments, causing stale execution whose irreversible effects persist even after the final call is corrected. We propose Repair-Aware Semantic Survival Gating (RASSG) to jointly control tool launch and cancellation during streaming speech. RASSG launches a tool when three consecutive speech recognition results arrive with identical nonempty arguments, then combines calibrated argument invalidity risk with execution costs to decide cancellation at the last feasible opportunity before effects become irreversible. Evaluation on ReCAST, FDB-v3, and Audio2Tool uses a controlled execution clock with fixed tool runtimes and cancellation parameters. On ReCAST's prospective-repair subset, RASSG reduces the stale execution rate from LocalAgreement-2's 50.00\% to 3.15\%, with a mean delay of 3.676~s across successful trajectories. The stale reduction also holds against every early-execution baseline on both public speech datasets. Ablations with identical initial launches support active cancellation, showing the value of controlling ongoing calls before their effects become irreversible. Code and data are available at \url{https://github.com/daijinhu/RASSG}.

\end{abstract}

\section{Introduction}
\label{sec:introduction}
Voice assistants are expanding from conversational responses to completing user tasks through external tools~\cite{pahwa2026audio2tool,zhang2026duplexsla,li2025dialogueagents,sui2026act}. Streaming speech processing lets systems interpret incoming speech incrementally and form the intents and arguments needed for tool use~\cite{jia2025efficient,chiang2026shanks}. Building on this capability, streaming tool execution allows operations to be prepared and launched before an utterance ends, so that speech understanding and task processing proceed in parallel~\cite{arora2025stream}. This overlap can shorten the time users wait for results and support more responsive, continuous interaction.

However, users may correct an argument before finishing their utterance~\cite{lin2026full}, leaving an already launched action inconsistent with their final intent. When an action using a superseded argument crosses its irreversible deadline, it causes \emph{stale execution}, even if the system eventually completes the correct call. Existing strategies face different limitations in handling this problem. Waiting for the complete utterance postpones work that could begin earlier. Recognition stability, as captured by LocalAgreement~\cite{liu2020low,machavcek2023turning}, supports early output but cannot guarantee that the user will not subsequently revise an argument. Waiting until a replacement argument is fully available before updating an action can miss the cancellation window. Streaming tool execution therefore requires joint consideration of whether the current argument is stable enough to launch and whether execution should continue while the tool remains cancellable.

To address the risk of stale execution when users revise arguments after tool launch, we propose Repair-Aware Semantic Survival Gating (RASSG). It coordinates early launch and repair handling in streaming speech tool use. RASSG launches tools based on observed argument stability and combines predicted argument invalidity with the tool's cancellation deadline to decide whether to cancel, allowing early execution to remain responsive to subsequent semantic changes.

\begin{figure}[t]
\centering
\includegraphics[width=\columnwidth,keepaspectratio]{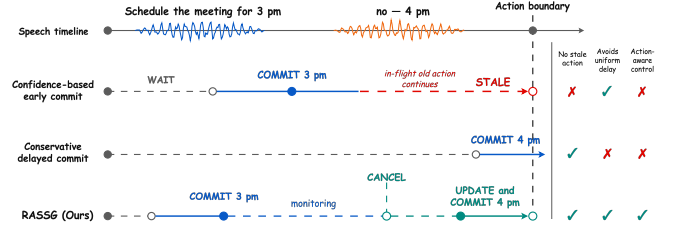}
\caption{An illustrative scheduling request changes from 3 pm to 4 pm. Confidence-based early commitment leaves the obsolete call running past its irreversible deadline (the action boundary), while conservative delayed commitment waits for the correction before launching. RASSG launches on stable arguments, monitors invalidity risk during execution, and cancels the old call before the deadline, then launches the revised request.}
\label{fig:rassg_example}
\end{figure}

We evaluate RASSG on synthetic speech with simulated tool execution. On the subset containing subsequent user repairs, RASSG reduces stale execution from 49.92\% to 3.15\% and successful delay from 4.434~s to 3.676~s relative to a control with identical initial launches but no active cancellation.

Our main contributions are as follows:
\begin{enumerate}
    \item We characterize execution correctness beyond final call success by jointly modeling argument versions and irreversible deadlines. This formulation captures stale execution that can coexist with eventual success and makes opportunities for repair during execution explicit.
    \item We introduce RASSG, which uses observed argument stability for launch and predicted argument invalidity together with cancellation deadlines for cancellation. This design lets the system act on stable input while continuing to assess whether an action should be withdrawn in response to subsequent repairs.
    \item We demonstrate reductions in stale execution and successful delay in controlled synthetic speech experiments relative to a control with identical initial launches and no active cancellation. Component and execution-semantics analyses identify the roles of launch stability, cancellation signals, and progress reuse, providing empirical evidence for the risk--latency tradeoff in streaming tool execution.
\end{enumerate}

\section{Related Work}
\label{sec:related}
Tool execution after complete input generates calls from a full utterance or a received dialogue turn. Speech-to-tool systems either map audio directly to calls or transcribe speech before parsing tools and arguments~\cite{pahwa2026audio2tool}. When requests lack necessary information, clarification methods recover unspecified intents and correct errors across turns~\cite{zhang2025asktoact}. Related benchmarks examine goal revisions, missing parameters, argument correction, and recovery from tool failures~\cite{alkhouli2025confetti,dong2025tool}. These developments improve the handling of incomplete or revised requests using available dialogue context. However, workflows that wait for the complete utterance leave tool processing until after the user finishes speaking.

Early execution from incremental input overlaps processing with ongoing speech. Incremental dialogue and streaming transcription provide mechanisms for revision propagation and stable-prefix confirmation~\cite{schlangen2011general,machavcek2023turning}. Related approaches move response generation and reasoning into the input period through speculative computation or semantic triggering~\cite{li2025predgen,zou2026lts}, while speech agents interleave or synchronize reasoning and tool actions with incoming audio~\cite{chiang2026shanks,zhang2026duplexsla}. Two works directly address changes to early calls. Stream RAG~\cite{arora2025stream} generates queries during speech; its model-triggered variant terminates earlier queries when a new one is issued. Speculative Interaction Agents~\cite{hooper2026speculative} handles user input and tool returns asynchronously, permits modification or removal of unfinished calls, and defers sensitive tool execution until confirmation. These mechanisms establish support for revising or withdrawing speculative calls. For tools that permit cancellation within a finite window, the remaining decision concerns whether an ongoing action should continue given future argument invalidity, cancellation cost, and the time remaining before irreversibility.

Evaluation has also expanded beyond isolated call accuracy to cover conversational correction and execution progress. Stateful and speech-based tool benchmarks collectively assess intermediate states, repairs in human speech, and verifiable task completion during full-duplex interaction~\cite{lu2025toolsandbox,lin2026full,ray2026tau,gurram2026evaluating}. These evaluations provide a basis for examining how evolving intent affects ongoing tool use. Within the incremental execution approach, RASSG focuses on cancellation decisions under finite tool deadlines. It uses observed argument stability for launch and predicted argument invalidity for cancellation, accounting for actual recognition arrivals and cancellation acknowledgements. This connects the decision to continue or withdraw an action with the joint evaluation of stale execution and successful delay.

\section{Methodology}
\label{sec:method}
\begin{figure*}[t]
\centering
\includegraphics[width=\textwidth]{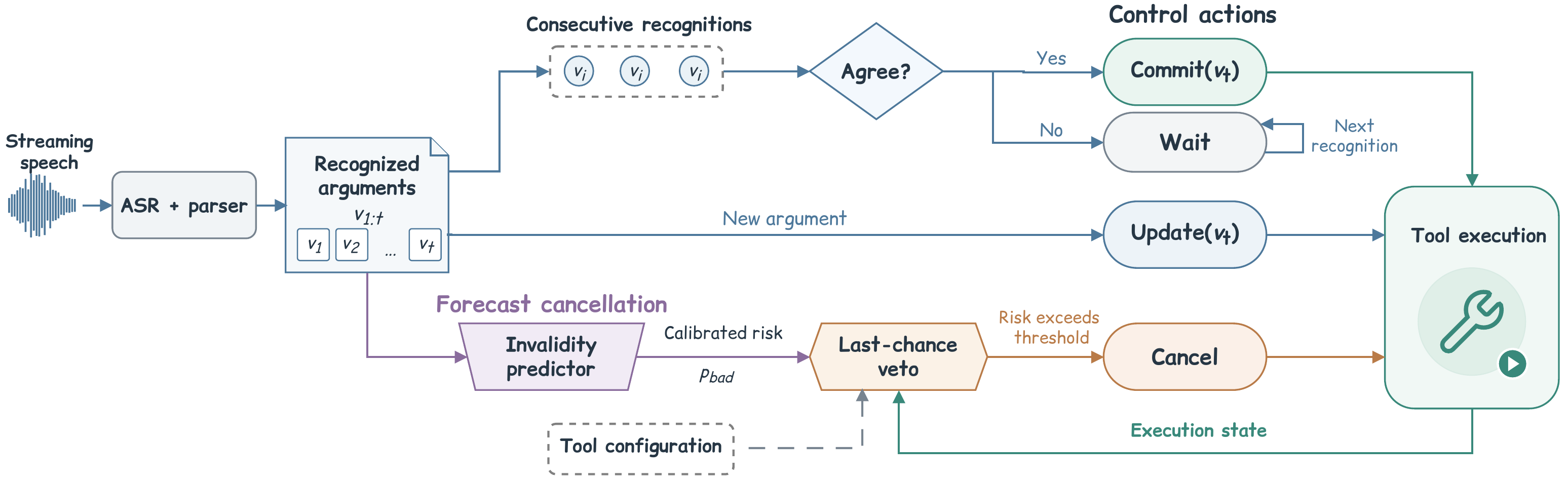}
\caption{RASSG control loop. Agreement across three actual ASR arrivals triggers launch. New arguments update active execution, while calibrated invalidity risk, tool configuration, and execution state determine cancellation at the last feasible opportunity.}
\label{fig:rassg_architecture}
\end{figure*}

RASSG coordinates early tool launch with repairs during execution. As shown in Fig.~\ref{fig:rassg_architecture}, parsed arguments feed stability-based launch and invalidity prediction, while newly recognized arguments update active calls. The controller combines calibrated risk with tool configuration and execution state to decide whether to cancel at the last feasible opportunity before the current version becomes irreversible.

\subsection{Task Formulation and Execution Clock}
At observation time $t$, the parser extracts a candidate argument $v_t$ from ASR results that have already arrived. Its history contains only observations available up to $t$. Transcription changes need not change the normalized argument. Different argument values define distinct execution versions. A version inconsistent with the final target causes stale execution if it survives its irreversible deadline, even when a later call succeeds. The final target is used only for offline supervision and scoring.

For tool $a$, we assume that runtime $\ell_a$, cancellation window $w_a$, and cancellation cost are fixed and known during an execution. Cancellation acknowledgement latency is also a fixed configuration parameter. A version launched at $c$ on a pipeline started at $s$ becomes irreversible at
\begin{equation}
 b=\min\{c+w_a,\ \max(c,s+\ell_a)\}.
 \label{eq:deadline}
\end{equation}
The deadline is bounded by both the cancellation window and completion time. A fresh call has $s=c$, while a timely argument update can retain the original pipeline start $s$. Progress reuse requires completed computation to remain valid after the argument changes, for example when preparation is independent of the revised field. Conveyor~\cite{xu2024conveyor} demonstrates tool interfaces that overlap such partial work with input generation; reuse across argument versions additionally requires that work to remain val

\subsection{Stability-Based Tool Launch}
Let $\hat v_i$ denote the normalized argument from the $i$th actual ASR arrival. An idle controller issues \COMMIT{} when the latest three arrivals satisfy $\hat v_{i-2}=\hat v_{i-1}=\hat v_i=v_t\ne\varnothing$; otherwise it issues \WAIT{}. This applies local agreement~\cite{liu2020low,machavcek2023turning} to argument values rather than complete transcriptions. Only a new ASR arrival changes the agreement history. Repeated control ticks using the same recognition therefore do not increase the agreement count. After the final ASR flush arrives, an idle controller can directly commit a nonempty candidate, ending regular waiting. Agreement provides evidence of current recognition stability, but a user may still revise that argument. Execution therefore remains subject to invalidity prediction and cancellation.

\subsection{Calibrated Argument Invalidity Prediction}
A causal GRU with 64 hidden units~\cite{cho2014learning} encodes confidence, parse history, prefix text, pauses, and version age. A supersession event is the earlier of the current version's annotated user-repair onset and the next nonempty ASR argument change. An unresolved repair denotes ongoing invalidity. The predicted distribution assigns masses $p_{0,t}$ to ongoing invalidity, $p_{k,t}$ to eight future intervals, $p_{>,t}$ to events beyond the prediction horizon, $p_{T,t}$ to a wrong version persisting to termination, and $p_{\emptyset,t}$ to a correct version persisting to termination. These outcome groups form a normalized distribution over the current argument version. The aggregate risk used for control is
\begin{equation}
 p_{\mathrm{bad},t}=p_{0,t}+\sum_{k=1}^{8}p_{k,t}+p_{>,t}+p_{T,t}.
 \label{eq:survival}
\end{equation}
This quantity estimates argument invalidity, with tool timing and costs entering the subsequent decision.

The training objective combines censored negative log-likelihood~\cite{lee2018deephit} and masked binary cross-entropy at cumulative event boundaries, with each argument version receiving equal total weight. A source-trained logistic calibrator rescales total supersession-event mass, preserving relative proportions within event and non-event branches and the conditional time shape. The probabilities in Eq.~\ref{eq:survival} denote the calibrated masses. Event annotations and final targets supply offline labels, whereas online prediction uses causal observations. The resulting aggregate risk drives cancellation. Launch remains governed by observed agreement. Conditional time shape is used by the separately evaluated lookahead-launch variant.

\subsection{Deadline-Aware Cancellation and Execution Updates}
For an active call whose argument still matches the nonempty candidate, the last-chance veto checks risk at the final regular decision that permits timely cancellation. Let $d_t$ include processing and any wait for previous acknowledgements, $\Delta$ be the control period, and $\kappa$ the cancellation acknowledgement delay. This opportunity satisfies $d_t+\kappa\le b<d_t+\Delta+\kappa$. Timeliness is determined by acknowledgement. Earlier decisions continue execution, allowing a replacement to arrive while progress remains reusable.

The veto compares the cost of continuing a potentially invalid version with the cost of cancelling and losing useful progress. Let $C_{\stale}$ and $C_{\cancel}$ denote stale-execution and cancellation costs, and let $\beta_a$ price delay. At decision time $d_t$, reusable progress is $g_t=\min\{\ell_a,\max(0,d_t-s)\}$. The controller cancels when
\begin{equation}
 C_{\stale}p_{\mathrm{bad},t}
 >C_{\cancel}+\beta_a g_t(1-p_{\mathrm{bad},t}).
 \label{eq:veto}
\end{equation}
The left side assigns stale-execution cost to predicted invalidity. The right side combines cancellation cost with the delay cost of discarding progress when the argument would remain valid. Thus, the cancellation threshold depends on both tool costs and accumulated execution progress. The rule uses no additional cost margin.

A new nonempty argument differing from the active version takes priority over the veto. If its \textsc{Update} is acknowledged by the current deadline, the replacement retains $s$ and receives a new version start and deadline under Eq.~\ref{eq:deadline}. A late replacement starts fresh execution and preserves any stale-execution record already incurred. \CANCEL{} instead clears the active call and its progress upon acknowledgement, without recommitting in the same observation. At a later observation, the idle controller may commit again once the launch condition holds. If the veto condition is not met, execution continues. Each transition updates the execution state used for the next decision.

\begin{table*}[!t]
\centering
\caption{Comparison of RASSG and baseline execution policies on ReCAST, FDB-v3, and Audio2Tool. Lower is better for all metrics.}
\label{tab:main_results}
\fontsize{9}{9.2}\selectfont
\setlength{\tabcolsep}{3pt}
\renewcommand{\arraystretch}{1.1}
\begin{tabular*}{\textwidth}{@{\extracolsep{\fill}}lrrrrrr@{}}
\toprule
 & \multicolumn{2}{c}{ReCAST} & \multicolumn{2}{c}{FDB-v3} & \multicolumn{2}{c}{Audio2Tool}\\
\cmidrule(lr){2-3}\cmidrule(lr){4-5}\cmidrule(lr){6-7}
Method & \shortstack{Stale execution\\rate (\%) $\downarrow$} & \shortstack{Successful\\delay (s) $\downarrow$} & \shortstack{Stale execution\\rate (\%) $\downarrow$} & \shortstack{Successful\\delay (s) $\downarrow$} & \shortstack{Stale execution\\rate (\%) $\downarrow$} & \shortstack{Successful\\delay (s) $\downarrow$}\\
\midrule
EOU & 0.58 & 8.676 & 40.00 & 23.880 & 52.73 & 9.196\\
LocalAgreement-2 & 50.00 & 3.954 & 65.00 & \textbf{9.210} & 62.91 & 8.135\\
Confidence commit & 21.08 & 5.633 & 73.00 & 9.507 & 73.92 & \textbf{7.910}\\
Stream RAG & 50.00 & 5.312 & 86.00 & 19.997 & 84.50 & 8.255\\
Speculative Tool Calling & 50.00 & \textbf{3.480} & 77.00 & 9.280 & 81.05 & 8.216\\
\midrule
\textbf{RASSG} & \textbf{3.15} & 3.676 & \textbf{57.00} & 11.072 & \textbf{54.65} & 8.797\\
\bottomrule
\end{tabular*}
\par\smallskip
\end{table*}
\section{Experiments}
\label{sec:setup}

\subsection{Dataset Design}
We construct ReCAST (Repair and Early Commitment Assessment for Streaming Tools), a synthetic speech dataset for evaluating early execution and subsequent repairs. Time, amount, location, and contact requests cover ordinary speech and different repair timings. Stable and corrected pairs share audio before branching. Source data contain 320 training and 128 development recordings. Speech-connected groups divide development equally into model-selection and policy-calibration sets. The test set contains 2,400 recordings from two new voices and new templates, with 1,200 prospective-repair recordings used for the main comparison.

Public evaluation covers all 100 FDB-v3 recordings~\cite{lin2026full}, and 2,145 Audio2Tool Tier~6 correction queries meeting the input-duration limit~\cite{pahwa2026audio2tool}, with one released rendition per query.

\subsection{Baselines}
We compare RASSG with five baselines: End-of-Utterance Execution, LocalAgreement-2, Confidence commit, and adaptations of Stream RAG and Speculative Tool Calling. End-of-Utterance Execution (EOU) waits for speech to end and final recognition to arrive. LocalAgreement-2~\cite{liu2020low,machavcek2023turning} launches after argument agreement across two actual ASR arrivals. Confidence commit~\cite{loth2018confidence} is our threshold control using source-selected confidence $0.95$. Stream RAG~\cite{arora2025stream} launches parallel calls for each arriving nonempty block proposal, including duplicates, then selects the earliest call matching the final observed plan and cancels unfinished alternatives at EOU. Speculative Tool Calling~\cite{hooper2026speculative} executes the first proposal immediately. Changed arguments trigger withdrawal and fresh execution after acknowledgement. Stream RAG assumes parallel tool capacity. Speculative Tool Calling evaluates eager execution, while its final-confirmation branch reduces to EOU under our atomic-plan representation.

\subsection{Evaluation Setup}
All policies share Whisper small.en~\cite{radford2023robust} prefixes decoded every 960~ms. ReCAST uses a fixed-lexicon parser. Public speech uses Qwen2.5-3B-Instruct~\cite{yang2025qwen3} with 4-bit weights to parse current prefixes and public tool schemas into complete plans. Arrivals include measured frontend service times. We execute tool calls within an event-clock execution model that processes commitment, update, cancellation acknowledgement, and irreversible-boundary events. Tool runtimes and cancellation costs are static configuration parameters. Control intervals, processing budgets, and acknowledgement delays are 320, 10, and 20~ms. Timely \textsc{Update} reuses progress. Restarting after \CANCEL{} does not.

We report two metrics. Stale execution rate is the fraction of all trajectories with at least one stale execution. Successful delay is measured only when the final argument is correct and an active call exists. Its origin is final-argument acoustic availability for ReCAST and waveform end for public speech, with negative values clipped to zero. Because these origins differ, delay comparisons are made within each dataset. Table delay means condition on each policy's own successful trajectories, whereas stale execution retains all trajectories.

\section{Results}
\label{sec:results}

\subsection{Main Results}
RASSG achieves the lowest stale execution rate among all four early-execution baselines on every dataset in Table~\ref{tab:main_results}. On ReCAST, it reduces stale execution by 93.71\% relative to LocalAgreement-2 while shortening successful delay by 0.278~s. It also removes 5.000~s of waiting relative to EOU. Against Speculative Tool Calling, stale execution falls from 50.00\% to 3.15\% for 0.197~s additional paired delay.

On public speech, RASSG lowers stale execution relative to Speculative Tool Calling by 20.00 percentage points on FDB-v3 and 26.39 on Audio2Tool. Reductions against Stream RAG are 29.00 and 29.84 points. All four early-execution comparisons on both public cohorts have stale-rate intervals excluding zero. With identical parsed proposals, these differences arise from execution control. Final selection among parallel calls or withdrawal after a replacement arrives can leave earlier effects irreversible; RASSG combines stronger launch evidence with a veto that can act before the parsed argument changes.

The high absolute stale rates on FDB-v3 and Audio2Tool largely reflect persistent plan errors in the shared ASR-to-tool parsing frontend. Of RASSG's stale trajectories, 70.18\% and 96.47\%, respectively, have a wrong nonempty final parser output and also incur stale execution under EOU. This exposure persists even when launch waits for complete input. On recordings with correct final parser outputs, RASSG lowers stale execution from LocalAgreement-2's 50.00\% to 32.14\% and from 20.10\% to 3.33\%. Thus, RASSG suppresses obsolete intermediate plans even when the frontend eventually resolves the request correctly.

\subsection{Ablation Study}
Table~\ref{tab:ablations} separates component removals from launch consensus. With first commitments and reactive argument updates fixed, removing active cancellation raises stale execution from 3.15\% to 49.92\%. Cancellation reduces stale execution by 46.77 percentage points, with a 95\% interval of [45.71, 47.78], and common-success delay by 0.758~s [0.702, 0.815]. The gain therefore comes from controlling an ongoing call, rather than delaying its first launch.
\begin{table}[!t]
\centering
\caption{Component ablations of RASSG on ReCAST, changing one design choice at a time.}
\label{tab:ablations}
\fontsize{9}{9.2}\selectfont
\setlength{\tabcolsep}{3pt}
\renewcommand{\arraystretch}{1.1}
\begin{tabular*}{\linewidth}{@{\extracolsep{\fill}}lrr@{}}
\toprule
Variant & \shortstack{Stale execution\\rate (\%) $\downarrow$} & \shortstack{Successful\\delay (s) $\downarrow$}\\
\midrule
\textbf{RASSG (full)} & 3.15 & 3.676\\
\midrule
\multicolumn{3}{l}{\textit{Component removals}}\\
Remove active cancellation & 49.92 & 4.434\\
Remove progress reuse & 3.06 & 4.510\\
\midrule
\multicolumn{3}{l}{\textit{Launch consensus}}\\
One-arrival consensus & 28.78 & 3.038\\
Two-arrival consensus & 16.86 & 3.274\\
Four-arrival consensus & 0.88 & 4.450\\
\bottomrule
\end{tabular*}
\end{table}

Removing progress reuse increases successful delay from 3.676 to 4.510~s, a paired penalty of 0.834~s [0.788, 0.881]. Its stale rate falls slightly, from 3.15\% to 3.06\%, because restarting can move the irreversible boundary later. The 0.083-percentage-point reduction comes from one recording across eight actions and three checkpoints, with a paired 95\% interval of [0.000, 0.256]. Progress reuse primarily saves waiting.

Consensus controls how much speech evidence precedes launch. Moving from two to three arrivals removes 13.71 percentage points of stale execution for 0.402~s additional waiting; moving from three to four removes another 2.27 points for 0.774~s. The fourth arrival costs more delay for a smaller absolute risk reduction. We adopt three-arrival consensus as the risk--delay operating point studied here and report the trade-offs of all other evaluated consensus settings.

\subsection{Results Analysis}

The method separates launch evidence from continuation risk. Argument agreement filters transient proposals before commitment; calibrated invalidity forecasts supersession after launch. Launch-only confidence or agreement rules do not make this second decision, and replacement-triggered withdrawal waits for the new argument. RASSG can act before replacement resolution, using the deadline and reusable progress to determine whether intervention remains feasible and its predicted benefit exceeds discarded work.

ReCAST's matched-launch trajectories directly support this mechanism. Among 28,800 recording--action--checkpoint combinations, cancellation removes stale execution from 13,470 trajectories without introducing a newly stale trajectory. In every avoided case, cancellation precedes the no-cancellation control's first irreversible error. Late replacements fall from 14,304 to 834: withdrawal followed by recommitment renews the version's cancellation window, allowing more subsequent corrections to arrive in time. Timely updates preserve valid work, explaining why stale reduction coincides with shorter delay. The reuse removal verifies this division of roles: the veto protects argument versions, while retained computation supplies the latency benefit.

\section{Conclusion}
\label{sec:conclusion}

RASSG combines argument-consensus launch with calibrated, deadline-aware cancellation. Under fixed execution parameters, it reduces stale execution relative to early-execution baselines across three speech datasets. Matched-launch removal experiments distinguish the protection supplied by cancellation from the waiting-time reduction supplied by progress reuse. Beyond three arrivals, further confirmation costs more waiting for less absolute risk reduction. Execution control must therefore remain responsive to repairs after launch.

\label{main-body-end}

\section{Compliance with Ethical Standards}
\label{sec:ethics}
The reported experiments use synthetic ReCAST speech and the publicly released FDB-v3 and Audio2Tool datasets~\cite{lin2026full,pahwa2026audio2tool}. They involve no participant recruitment, collection of new human recordings, or intervention with human participants.

\section{Acknowledgments}
\label{sec:acknowledgments}
No funding was received for this study. There are no conflicts of interest, financial or non-financial.

Large language models were used for (i) assistance with manuscript drafting and polishing, (ii) related-work search and discovery, and (iii) code generation.

\clearpage

\bibliographystyle{IEEEbib}
\bibliography{strings,refs}

\end{document}